%% file: DoNet.tex
\documentclass[10pt,twocolumn,letterpaper]{article}

\usepackage{cvpr}  
\usepackage{graphicx}
\usepackage{amsmath}
\usepackage{amssymb}
\usepackage{booktabs}
\usepackage[pagebackref,breaklinks,colorlinks]{hyperref}
\usepackage[capitalize]{cleveref}
\usepackage{color}
\crefname{section}{Sec.}{Secs.}
\Crefname{section}{Section}{Sections}
\Crefname{table}{Table}{Tables}
\crefname{table}{Tab.}{Tabs.}
\usepackage{multirow}
  
\usepackage[accsupp]{axessibility}  % Improves PDF readability for those with disabilities.

\def\confName{CVPR}
\def\confYear{2023}
\definecolor{yn}{RGB}{144, 180, 75}
\begin{document}

%%%%%%%%% TITLE
\title{DoNet: Deep De-overlapping Network for Cytology Instance Segmentation}

\author{
 Hao Jiang$^1$\thanks{Equal contribution} \hspace{1.0cm} 
 Rushan Zhang$^{1}$\footnotemark[1] \hspace{1.0cm} 
 Yanning Zhou$^2$\hspace{1.0cm} 
 Yumeng Wang$^{1}$\hspace{1.0cm} 
 Hao Chen$^{1}$\thanks{Corresponding author}\hspace{1.0cm}\\
 $^1$The Hong Kong University of Science and Technology \hspace{1.0cm} 
 $^2$Tencent AI Lab\\
 \texttt{\footnotesize \{hjiangaz,jhc\}@cse.ust.hk, \{rzhangbq,ywanglu\}@connect.ust.hk, amandayzhou@tencent.com}}
\maketitle

%%%%%%%%% ABSTRACT
\begin{abstract}
    Cell instance segmentation in cytology images has significant importance for biology analysis and cancer screening, while  remains challenging due to 1) the extensive overlapping translucent cell clusters that cause the ambiguous boundaries, and 2) the confusion of mimics and debris as nuclei.
    In this work, we proposed a \textbf{D}e-\textbf{o}verlapping \textbf{Net}work (DoNet) in a decompose-and-recombined strategy.
    A Dual-path Region Segmentation Module (DRM) explicitly decomposes the cell clusters into intersection and complement regions, followed by a Semantic Consistency-guided Recombination Module (CRM) for integration.
    To further introduce the containment relationship of the nucleus in the cytoplasm, we design a Mask-guided Region Proposal Strategy (MRP) that integrates the cell attention maps for inner-cell instance prediction.
    We validate the proposed approach on ISBI2014 and CPS datasets. Experiments show that our proposed DoNet significantly outperforms other state-of-the-art (SOTA) cell instance segmentation methods.
    The code is available at ~\url{https://github.com/DeepDoNet/DoNet}.
\end{abstract}

%%%%%%%%% Intro
\section{Introduction}
\input{S1_Introduction.tex}

%%%%%%%%% Related work
\section{Related Work}
\input{S2_Related_Work.tex}

%%%%%%%%% Method
\section{Methodology}
\input{S3_Method.tex}

%%%%%%%%% Experiments
\section{Experiments}
\input{S4_Experiment.tex}

%%%%%%%%% Conclusion
\section{Conclusion}
\input{S5_Conclusion.tex}

%%%%%%%%% Acknowledgement
\section*{Acknowledgement}
\input{S6_Acknowledgement.tex}

\newpage
%%%%%%%%% REFERENCES
{\small
\bibliographystyle{ieee_fullname}
\bibliography{egbib}
}

\end{document}

%% file: S1_Introduction.tex
\label{sec:intro}
\begin{figure}[!ht]
  \centering
   \includegraphics[width=0.85\linewidth]{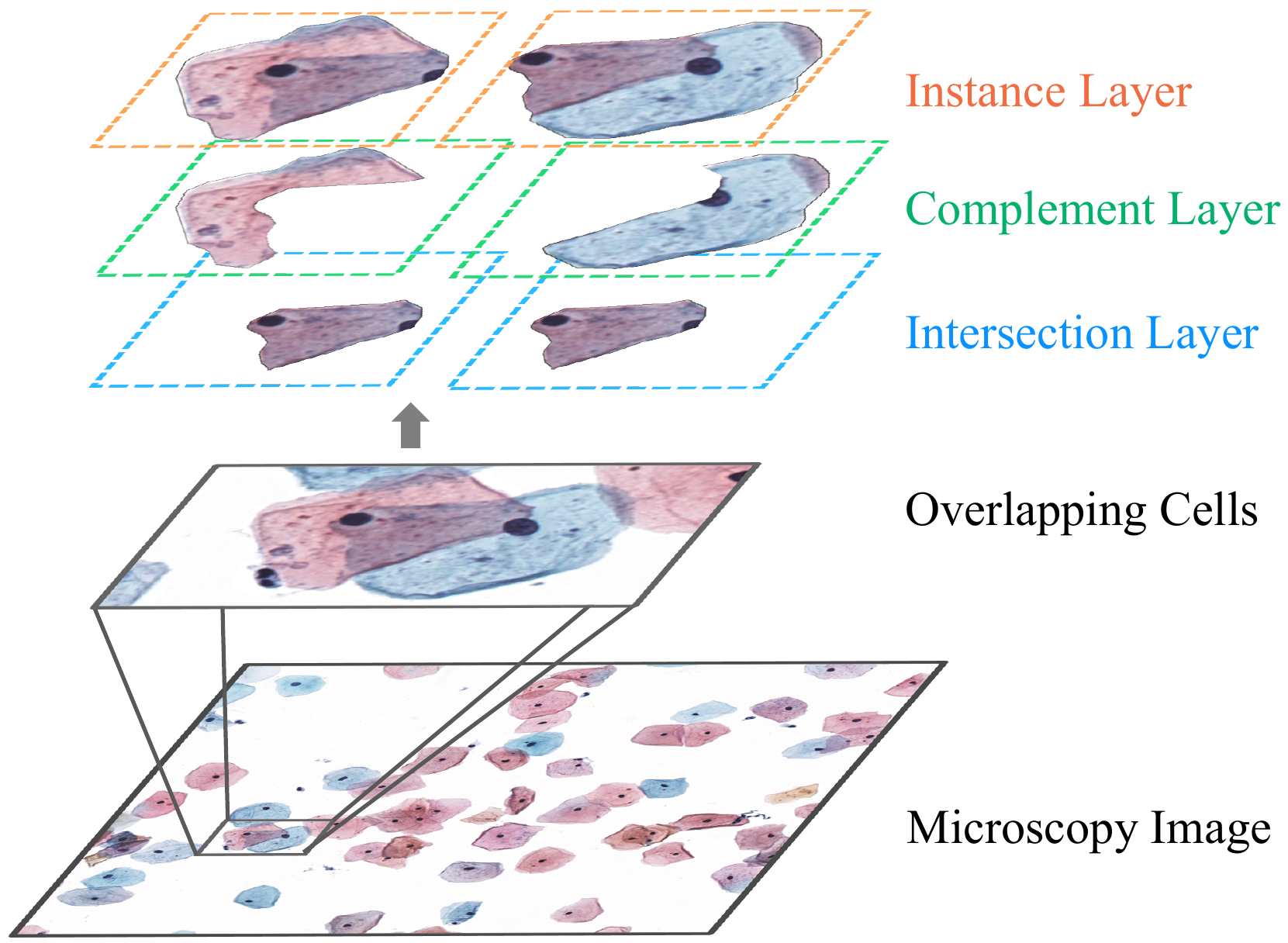}
   \caption{The schematic illustration of proposed DoNet with decompose-and-recombined strategy, which maps each overlapping cell into the intersection, complement, and instance layers to address the overlapping issue in cytology instance segmentation.}
   \label{fig1}
\end{figure}

Cytology image has been essential for cancer screening and earlier diagnosis, such as qualitative and quantitative identification of cellular morphology, nuclei size, nuclear-cytoplasmic ratio, and other cytological features \cite{hoda2007fundamentals,nayar2015bethesda,jiang2020geometry}. However, examining tens of thousands of cells under the microscope visually is inherently tedious and suffers from inter-/intra-observer variability. Computational techniques enable efficient and accurate characterization of cells from cytology images \cite{hoda2007fundamentals,jiang2022deep}. Among all computational techniques, cell segmentation has been a fundamental and widely-studied task, since the acquisition of cell-level identification is a pre-requisition for further assessment and analysis\cite{lin2021dual,chai2022deep}. 

Deep Learning (DL) methods show promising results for cell-nuclei segmentation in the histopathology image\cite{hussain2020shape,chen2017dcan,jin2023labelefficient,chen2016dcan}. However, cytology segmentation remains challenging for the following two reasons. \textbf{Firstly}, cells in a cytology image are prone to cluster with each other, leading to the overlapping issue. In the cytology images, the translucent cytoplasm of the cell (seen in Figure \ref{fig1}) tends to occlude each other with low contrast staining, leading to ambiguous cellular boundary predictions. This phenomenon is particularly evident in cervical cell images. \textbf{Secondly}, hard mimics, are widespread in the background, along with other technical artefacts such as bubbles, which could mislead the instance segmentation models \cite{che2022learning}. Take the cervical cell image as an example, the widespread white blood cells and mucus stains lead to false predictions for nuclei. To address these challenges, several works\cite{lu2015improved, ushizima2015segmentation} propose the segment-then-refine paradigm, while others \cite{zhou2019irnet,zhou2020deep} utilize the detection-based framework, e.g., Mask R-CNN \cite{he2017mask}. However, they fail to model the interaction between intersection and complement sub-regions within the translucent cell cluster explicitly, resulting in a limited understanding of cross-region relationships.

Amodal instance segmentation tackles the occlusion problem by inferring the integral object based on the partially visible region  \cite{li2016amodal}. Based on the fact that humans can infer the occluded region of an object despite the ambiguity, these methods attempt to learn the integrated object mask (amodal mask) for better occlusion reasoning capability\cite{follmann2019learning,xiao2021amodal} via synthesizing occluded data label pairs and aggregating global information to enhance perceptual ability.
Compared to natural scenes, cell instances in cytology images are mostly semi-transparent. Therefore, an occlusion (overlapping) region exits in both the occluding and occluded instances. However, treating semi-transparent overlapping regions as general occlusion regions is not optimal, since they have different appearances compared to non-overlapping regions, and could in fact provides richer shape information than general occlusion regions.

Motivated by the amodal perception, we propose a decompose-and-recombine strategy for translucent cell instance segmentation, named De-overlapping Network (DoNet). Figure \ref{fig1} provides the schematic diagram. For each cell cluster with more than one cellular sub-region, DoNet starts from implicitly learning the hidden interaction of sub-regions by predicting instance masks from clusters. Then, it explicitly models the components and their relationships via the intersection layer, complement layer, and instance layer, to enhance its perceptual capability.

Initially, we adopt Mask R-CNN to get the coarse predictions, followed by a novel Dual-path Region segmentation Module (DRM) that combines features and coarse masks from the first stage to decompose cell clusters into intersection and complement sub-regions. Then, the semantic Consistency-guided Recombination Module (CRM) is designed to encourage consistency between the refined instances and integral sub-region predictions. Furthermore, to impose the morphological constraint that nuclei stay inside the cellular regions, we propose a Mask-guided Region Proposal Module (MRP) to encourage the model to focus on the intra-cellular area during nuclei segmentation.

The overall contributions are summarized as follows:
\begin{itemize}

\item A novel de-overlapping network for cell instance segmentation with a decompose-and-recombined strategy, decomposing the cell regions with the DRM, as well as implicitly and explicitly modeling the semantic relationship between intersection, complement, and instance (cell) components via the CRM.
These designs equip the network with enhanced perceptual capability in overlapping cellular sub-regions.  

\item A mask-guided region proposal module (MRP) that leverages the cytoplasm attention map for the intra-cellular nuclei refinement, which imposes the biology prior of cellular instances into the module, effectively mitigating the influence of mimickers widespread in the background.

\item Extensive experiments on two overlapping cytology image segmentation datasets, namely ISBI2014 \cite{lu2015improved} and CPS \cite{zhou2020deep}, demonstrating that our proposed DoNet outperforms other state-of-the-art (SOTA) methods by a large margin.   
\end{itemize}

\begin{figure*}[t]
\centering
\includegraphics[width=\textwidth]{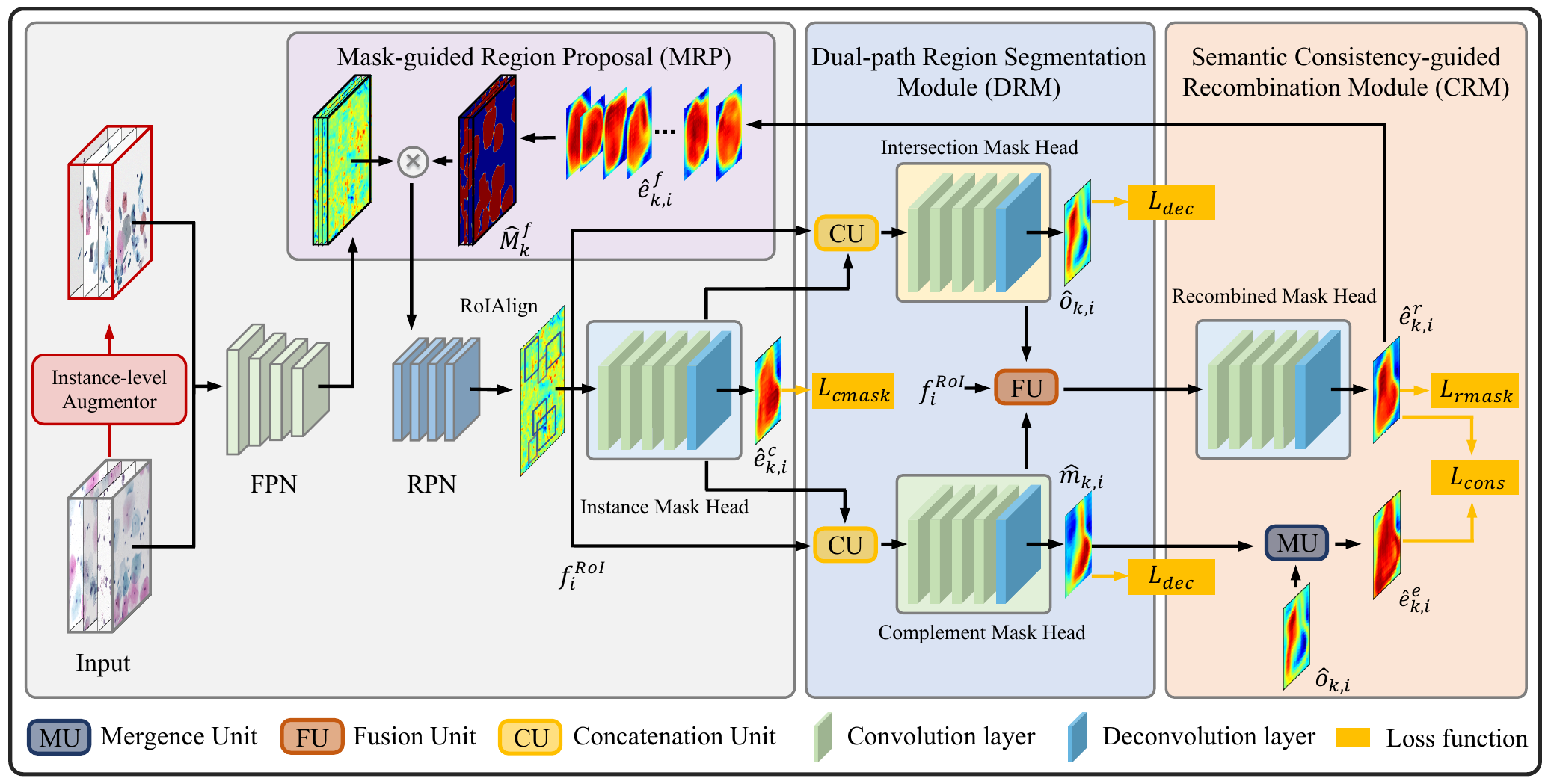} 
\caption{The flowchart of the proposed DoNet. 
It consists of four main parts: (1) baseline for coarse mask segmentation; (2) DRM to simultaneously regress intersection and complement regions which provides cues for final mask refinement; (3) CRM to explicitly refine the final mask and encourage the semantic consistency; (4) MRP to mitigate the effect of background noise.}
\label{fig2}
\end{figure*}

%% file: S2_Related_Work.tex
\label{sec:rela}

\noindent \textbf{Cytology Instance Segmentation:}
Challenges such as cell clustering, hard mimics, and semi-transparent overlapping cytoplasm regions pose in the segmentation of cellular instance (e.g., cell, nuclei, cytoplasm) from cytology images. These challenges inspired numerous insightful works, especially after the publication of ISBI2014\cite{ushizima2015segmentation}. Two mainstream approaches are proposed to tackle the overlapping cell segmentation challenge: the segment-then-refine stream and the end-to-end training stream.

Early approaches tackling this challenge are primarily the combination of pixel-level segmentation models and additional post-processing techniques, such as seeded watershed algorithm \cite{kowal2020cell}, random walk \cite{zhang2020polar}, conditional random field algorithms \cite{wan2019accurate}, and star-convex parameterization \cite{walter2021multistar}. To further improve the segmentation performance, some other methods introduce morphological prior. Song et al. design a dynamic multi-template deformation model, which leverages case-specific shape constraints to guide the inference of overlapping cell boundaries\cite{song2016accurate}. Tareef et al. propose to refine cytoplasm boundary via a learnable shape prior, which is dynamically modeled as the fusion of shape templates \cite{tareef2017optimizing}. 

Escaping from the complex post-processing procedure, many later works turn to end-to-end training. These approaches typically adopt Mask R-CNN as their baseline. The first attempt following this approach was \cite{sompawong2019automated}, a model for nuclear instance segmentation from liquid-based cytology smears. Based on the appearance similarity between different cells, Zhou et al. propose the IRNet to explore instance relation during overlapping cervical cell segmentation \cite{zhou2019irnet}. To leverage information from unlabeled data, Zhou et al. provide a mean-teacher-based semi-supervised learning scheme, MMT-PSM \cite{zhou2020deep}. 

The above two methods have achieved notable performance improvements, yet still lack the necessary perceptual capability for overlapping regions, leading to sub-optimal results, i.e., ambiguous cell boundaries. In this study, we propose a de-overlapping network, named DoNet, to implicitly and explicitly model the interaction between the integral instance and its sub-regions. 

\noindent \textbf{Occluded Instance Segmentation:}
Instance segmentation, a typical task in computer vision, refers to the inference of bounding boxes and instance segmentation masks. Most notable approaches follow the detect-then-segment paradigm, such as Mask R-CNN \cite{he2017mask} and its following variants \cite{huang2019mask,cai2018cascade}. However, these approaches could not handle the occlusion problem, common in visual application scenarios like robotic manipulations, scene parsing, and autonomous driving \cite{back2022unseen,geiger2012we}. Due to the inconsistencies within a Regions of Interest (RoI), these models suffer from the limited perceptual capability, leading to ambiguous segmentation boundaries. To tackle this issue, a specific task, occluded instance segmentation aims at leveraging visible regions to  perceive the entirety of the occluded instance accurately. This capability of inferring and reasoning occluded objects is defined as amodal perception \cite{follmann2019learning}, thus occluded instance segmentation is also called amodal instance segmentation.

Research on amodal instance segmentation starts from \cite{li2016amodal}, which poses this challenging task and gives a synthesis-based solution. In this work, the first amodal instance segmentation dataset, the COCOA dataset, is established based on the COCO dataset. Afterward, some notable solutions accomplish this task and achieve state-of-the-art performance. Previous approaches tackle this challenge by leveraging information from visible instance segmentation, then inferring the amodal mask \cite{follmann2019learning,back2022unseen}. For example, Occlusion R-CNN (ORCNN) is built on R-CNN with two mask heads, the visible mask head, and the amodal mask head, which directly predict the amodal mask and the visible mask, inferring the visible mask through the subtraction of these two masks \cite{follmann2019learning}. Qi et al. \cite{qi2019amodal} perform amodal instance segmentation with multi-level coding and establish a large-scale amodal instance segmentation dataset. Xiao et al. \cite{xiao2021amodal} leverage shape prior knowledge to infer amodal mask using memory codebook. Some other models, like BCNet \cite{ke2021deep} abandon the visible instance prediction based manner and directly models the relationship between the occluder and the occludee. To address the lack of amodal mask ground truth, ASBU \cite{nguyen2021weakly} introduces a weakly supervised amodal segmenter, which generates pseudo-ground truth by boundary uncertainty estimation.

Essentially, the above occluded instance segmentation methods aim to enhance the amodal perception and reasoning capability, inspiring us to explore the interaction between instance sub-regions and the integral instance. Compared to amodal segmentation, whose focus on inferring the invisible region from the partially visible regions, our focus is to resolve the inconsistency between the intersection and complement regions.

%% file: S3_Method.tex
\subsection{Overview}\label{Overview}

\noindent \textbf{Problem formulation:} Following the setting of cell instance segmentation, we are provided with a dataset with $K$ images and the corresponding annotations $\mathcal{D} =  \left\{(\mathcal{X}_{k},\mathcal{Y}_{k})\right\}_{k=1}^{K}$, each of which contains annotations of bounding boxes $\mathcal{B}_{k}=\left\{b_{k, i}\right\}^{N_{k}}_{i=1}$, object categories  $\mathcal{C}_{k} = \left\{c_{k, i} \right\}^{N_{k}}_{i=1}$, and instance masks  $\mathcal{E}_{k} = \left\{e_{k, i}\right\}^{N_{k}}_{i=1}$, where $N_{k}$ denotes the number of instances in the $k$-th image.
Here, we focus on segmenting the nucleus and cytoplasm for cell instance $c_{k, i} \in \left\{\text{nuclei}, \text{cytoplasm} \right\}$. For each cell cluster, based on the localization relationship among the cells, we decompose the instance mask into the intersection region $\mathcal{O}_{k} = \left\{o_{k, i}\right\}^{N_{k}}_{i=1}$ and the complement region
$\mathcal{M}_{k} = \left\{m_{k,i}\right\}^{N_{k}}_{i=1}$ via the logical operation (see in Figure \ref{fig1}), which is used to model their relationship in the DoNet.

\noindent \textbf{Workflow:} 
The flowchart of the proposed DoNet is provided in Figure \ref{fig2}. It takes Mask R-CNN \cite{he2017mask} as the base model, followed by a Dual-path Region segmentation Module (DRM) which takes the instance features as input to predict intersection regions $\hat{o}_{k, i}$ and complement regions $\hat{m}_{k, i}$ of cytoplasm via the intersection and complement layers (Section \ref{DRM}). 
After decomposing cell clusters according to the position relationship, a Consistency-guided Recombination Module (CRM) takes instance features from DRM and RoIAlign layer to generate the recombined masks $\hat{e}^{r}_{k, i}$ for consistency regularization (Section \ref{DRM}).
Then, the recombined cytoplasm prediction is utilized as prior knowledge for intra-cellular object prediction throughout the Mask-guided Region Proposal (MRP) (Section \ref{MRP}).
We will present the details of our framework in the following sections.

\noindent\textbf{Coarse Mask Segmentation:} Previous research works on Mask R-CNN has shown competitive performance in instance segmentation \cite{he2017mask, huang2019mask, follmann2019learning}, adopted into the proposed DoNet for coarse mask segmentation. Mask R-CNN consists of two stages. The first stage utilizes Feature Pyramid Network (FPN) for feature extraction and a Region Proposal Network (RPN) for candidate object bounding boxes generation. The second stage generates Region-of-Interest (RoI) features $f^{roi}_{k, i}$ via the RoIAlign layer and predicts object classes $\hat{c}_{k, i}$ and bounding boxes $\hat{b}_{k, i}$ from the detection head, and semantic masks $\hat{e}^{c}_{k, i}$ from the Instance Mask Head ($H_{i}$), respectively. Due to limited perception capability in overlapping regions, $\hat{e}^{c}_{k, i}$ may contain ambiguous boundaries. Thus, we denote $\hat{e}^{c}_{k, i}$ as the coarse mask, which provides information for sub-region decomposition in DRM and suppresses the interference from background mimics. Following standard losses in \cite{he2017mask}, a multi-task loss $\mathcal{L}_{coarse}$ for coarse mask segmentation is formed as,
\begin{equation}
  \mathcal{L}_{coarse} = \mathcal{L}_{reg} + \mathcal{L}_{cls} + \mathcal{L}_{cmask},
  \label{eq:eq1}
\end{equation}
where $\mathcal{L}_{reg}$ denotes the smooth-L1 loss for bounding box regression, $\mathcal{L}_{cls}$ denotes cross-entropy (CE) loss for classification, and $\mathcal{L}_{cmask}$ denotes pixel-wise CE loss for segmentation.

\subsection{Decompose-and-recombined Strategy} \label{DRM}
Previous solutions for amodal perception infer the integral structure of occluded instances using visible regions\cite{follmann2019learning,xiao2021amodal}. However, they ignore information from overlapping regions, which tend to be sub-optimal in the case of complex superposition of translucent objects. To empower the perception capability of model for overlapping translucent regions, we specifically design the decompose-and-recombined strategy.

\noindent\textbf{Dual-path Region Segmentation Module (DRM):}
As shown in Figure \ref{fig2}, DRM consists of an Intersection Mask Head $H_{o}$ and a Complement Mask Head $H_{m}$ with the same architecture. Let $f_{k, i}^{c}$ denotes the rich semantic feature before the coarse mask prediction in the Instance Mask Head. Both $H_{o}$ and $H_{m}$ take the concatenation of $f^{roi}_{k, i}$ and  $f_{k, i}^{c}$ as input to predict the intersection and complement regions $\hat{o}_{k, i}, \hat{m}_{k, i}$ in clusters. See the supplementary material for details of the Concatenation Unit. Specifically, each head consists of 4 convolutional layers to generate features in $14 \times  14 \times 256$, followed by one deconvolutional layer to get the semantic mask with a resolution of $28 \times 28 \times 1$. Here, we add the pixel-wise CE loss to both heads as the explicit constraint for decomposition,
\begin{equation}
\begin{aligned}
  \mathcal{L}_{dec} = \frac{1}{K}\sum_{k=1}^{K}\frac{1}{N_{k}}\sum_{i=1}^{N_{k}} \left ( \mathcal{L}_{ce} (\hat{o}_{k, i}, o_{k,i})
  + \mathcal{L}_{ce}(\hat{m}_{k, i}, m_{k,i})\right ).
  \label{eq:eq4}
\end{aligned}
\end{equation}

\noindent\textbf{Semantic Consistency-guided Recombination Module (CRM):}
To further enhance the perception capability for overlapping instances, CRM is designed to encourage DoNet to perceive integral instances. Specifically, let $f_{k,i}^{o}$ and $f_{k,i}^{m}$ denote the features before the last layer in the intersection mask head and the complement mask head. The rich semantic feature for the overlapping and non-overlapping region is considered as the residual information for the complex areas which is fused with $f^{roi}_{k, i}$ and then fed into the recombined mask head. See the supplementary material for details of the Fusion Unit. Noted that we reuse the Instance Mask Head ($H_{i}$) here to predict the integral instance. This combination facilitates CRM to leverage contextual information of overlapping instances, leading to the improvement of perception capability. The refined mask $\hat{e}_{k,i}^{r}$ from CRM is optimized by segmentation loss $\mathcal{L}_{rmask}$,
\begin{equation}
  \mathcal{L}_{rmask} = \frac{1}{K} \sum_{k=1}^{K}\frac{1}{N_{k}} \sum_{i=1}^{N_{k}}{\mathcal{L}_{ce}(\hat{e}_{k,i}^{r}, e_{k,i}) }.
  \label{eq:eq5}
\end{equation}

To further enhance the feature representation capacity for the relationship of components, we add semantic consistency regularization between the recombination $\hat{e}_{k, i}^{r}$ and merged predictions of $\hat{o}_{k, i}$ and $\hat{m}_{k, i}$,
\begin{equation}
  \mathcal{L}_{cons} = \frac{1}{K}\sum_{k=1}^{K} \frac{1}{N_{k}} \sum_{i=1}^{N_{k}}{\mathcal{L}_{ce}\left (\hat{e}_{k,i}^{r}, \mathcal{F}_{merge}\left (\hat{o}_{k, i}, \hat{m}_{k, i}\right )\right ) },
  \label{eq:eq6}
\end{equation}
where $\mathcal{F}_{merge}( \cdot)$ represents the merging operation for the intersection region and the complement region in Mergence Unit, which is $xor()$, two sub-region masks are passed through Sigmoid function for normalization. Then, we calculate the Mask Exclusive-OR of two mask logits for merging them and suppressing redundant predicted pixels.

\subsection{Mask-guided Region Proposal} \label{MRP}
It is inevitable for cytology images to exist abundant cellular debris and non-target objects, e.g., white blood cells and mucus. These objects often have a similar appearance as the nuclei, with small round surfaces stained in dark purple, which increases the model identification difficulty. To avoid the interference it brings,  we further propose a Mask-guided Region Proposal module (MRP) to encourage the model to generate nuclei proposals in intra-cellular regions.

For each image, we can aggregate all the recombined instance predictions $\hat{e}_{k, i}^{r}$ in CRM together into a semantic mask $\hat{M}_{k}$. The raw predictions are first mapped back and then summed up according to their bounding boxes predictions $\hat{b}_{k, i}$ in the detection head, followed by a Sigmoid function to normalize them into probabilities. $\hat{M}_{k}$ is considered as the attention score to re-weight features $f_{k}$ in origin FPN via element-wise multiplication:
\begin{equation}
  f_{k}^{w} = \hat{M}_{k} \circ f_{k},
  \label{eq:eq2}
\end{equation}
where $f_{k}^{w} $ denotes the re-weighted features for nuclei proposal predictions in MRP. Thus, extracellular pixels with a lower probability to be cytoplasm are suppressed, which reduces the false positives for background instances (i.e., blood, mucus, and others). In addition, MRP also builds information interaction between different stages, which naturally encourages feature representation ability.

\subsection{End-to-end Learning}\label{E2e}
\noindent \textbf{Extended dataset with synthetic clusters:}
The semi-transparency characteristic of cytoplasm alleviates the effect of high overlap, allowing cytologists to delineate the contour of cellular instances based on expert knowledge. However, extremely challenging labeling and unavoidable label noise limit the availability of large-scale annotated datasets. To tackle this problem, we propose an instance-level data augmentor for overlapping cell data augmentation, which can generate large-scale synthetic cell clusters  with controllable overlapping ratios and transparency based on the annotated cellular instances. This synthetic dataset further facilitates the generalization ability, by providing more diverse data to implicitly learn the concept of instance and its components. See the supplementary material for details of the synthetic pipeline. Unless otherwise specified, we do not use this synthetic dataset in the following comparison.

\noindent \textbf{Overall Loss Function:} 
The proposed cell instance segmentation framework can be trained in a supervised manner. The overall objective $\mathcal{L}$ is as follows,
\begin{equation}
\mathcal{L} =  \mathcal{L}_{coarse} + \lambda_{dec}\mathcal{L}_{dec} + \lambda_{rmask}\mathcal{L}_{rmask} + \lambda_{cons}\mathcal{L}_{cons},   
  % \mathcal{L}_{Sup} = \mathcal{L}_{co} + \lambda_{DRM} \mathcal{L}_{DRM} + \lambda_{CRM} \mathcal{L}_{CRM}
  \label{eq:eq8}
\end{equation}
where $\mathcal{L}_{coarse}$ denotes losses for RoI extraction and coarse mask prediction, $\mathcal{L}_{dec}$ denotes the decomposition loss for intersection and complement regions segmentation, $\mathcal{L}_{rmask}$ is the segmentation loss for refined masks, $\mathcal{L}_{cons}$ supervises the semantic consistency between integral instance and sub-regions. $\lambda_{dec}$, $\lambda_{rmask}$, and $\lambda_{cons}$ are trade-off parameters controlling the importance of each component.

%% file: S4_Experiment.tex
\subsection{Experimental Setup}
We evaluate our proposed DoNet in two cytology image datasets for overlapping cell instance segmentation:

\noindent \textbf{ISBI2014\cite{lu2015improved}:} This is a widely-used dataset from \emph{Overlapping Cervical Cytology Image Segmentation Challenge}\footnote{~\url{https://cs.adelaide.edu.au/~carneiro/isbi14_challenge/index.html}}, which consists of 8 extended depth-of-focus (EDF) real cervical cytology images and corresponding synthetic images. It contains high-quality pixel-level annotations for both nuclei and cytoplasm with a resolution of $512 \times 512$. We follow the setting in this challenge \cite{lu2015improved} to use 45, 90 and 810 images for training, validation and testing to evaluate our proposed DoNet in a supervised setting.

\noindent \textbf{CPS\cite{zhou2020deep}:} This liquid-based cytology dataset contains 137 labeled images with a resolution of $1000 \times 1000$. In total, it contains 4439 cytoplasm and 4789 nuclei annotations. We conduct 3-fold cross-validation on this dataset.

\noindent \textbf{Evaluation metrics:} To measure the overall performance of the proposed DoNet, we utilize four commonly-used evaluation metrics in instance segmentation: aggregated Jaccard index (AJI), average Dice coefficient (Dice), F1-score (F1)\cite{kumar2017dataset}, mean of Average Precision (mAP) \cite{he2017mask}. In order to compare our result on the ISBI2014 with previous studies \cite{lu2015improved}, we further adopt evaluation metrics including Dice, object-based false negative rate ($\text{FN}_{o}$), and pixel-based true positive rate ($\text{TP}_{p}$).

\input{Table/T1.tex}

\noindent \textbf{Implementation details:}
We utilize the Mask R-CNN\cite{he2017mask} in Detectron2 \cite{wu2019detectron2} as the baseline model.
We use the ResNet-50-based FPN network in all experiments.
During training, we adopt SGD with 0.9 momentum as the optimizer. 
We set the initial learning rate to 0.001 and add the linear warm-up in the first 1k iterations. 
We train the network for 60k iterations, decreasing the learning rate by a factor of 0.1 after 50k and 55k iterations.

\subsection{Results}
We quantitatively compare the cell instance segmentation results from the ISBI2014 and CPS in Table \ref{t1} with state-of-the-art methods in the field of general instance segmentation (Mask R-CNN\cite{he2017mask}, Cascade R-CNN\cite{cai2018cascade}, Mask Scoring R-CNN\cite{huang2019mask}, HTC\cite{chen2019hybrid}) and amodal instance segmentation (Occlusion R-CNN\cite{follmann2019learning}, Xiao et al. \cite{xiao2021amodal}). Noted that the amodal instance segmentation methods and general instance segmentation methods perform differently on two datasets due to the varying degrees of overlapping. Our method achieves the highest scores among all metrics. Specifically, it gains $2.68\%$ and  $0.52\%$ improvements for mAP and AJI compared with the best of others\cite{follmann2019learning} on the ISBI2014 dataset, as well as $1.85\%$ and $1.02\%$ improvements compared with the best\cite{xiao2021amodal} on the CPS dataset. We also evaluate $TP_{p}$ and $FN_{o}$ for DoNet to compare results against winners of the ISBI2014 challenge\cite{nosrati2014variational, ushizima2015segmentation} and their following works \cite{lee2016segmentation, tareef2018multi}, which are mostly the segment-then-refine manners
(Seen in Table \ref{t2}).

Furthermore, by introducing the synthetic clusters as instance-level augmentation, DoNet further has $0.45\%$ and $0.68\%$ improvements for mAP and AJI on the CPS dataset.This is mainly because the synthetic overlapping cells can further enhance the model's occlusion reasoning capability, which is consistent with the conclusions in \cite{li2016amodal}.

\input{./Table/T2.tex}

\subsection{Ablation Study}
\noindent \textbf{Effects of Network Components:} 
We perform ablation studies to investigate the effects of the different components of our proposed pipeline for DoNet. The comparison results can be seen in Table \ref{t3}. By adding DRM for explicit decomposing integral instances into intersection and complement sub-regions, we observe a $3.25\%$ increase in the average mAP of cytoplasm and nuclei on ISBI2014. However, directly adding DRM without the fusion of structural and morphological information, may mislead the model. This issue is observed in the more complex CPS dataset. To alleviate this problem, DoNet takes advantage of the decompose-and-recombined strategy by adding CRM after DRM.This strategy brings a total of $7.34\%$ and $1.74\%$ mAP improvements on two datasets by strengthening the model's perception of overlapping regions while preserving morphological information.

\begin{figure}[!ht]
% \vspace{-0.05in}
	\centering
	\includegraphics[width=0.95\linewidth]{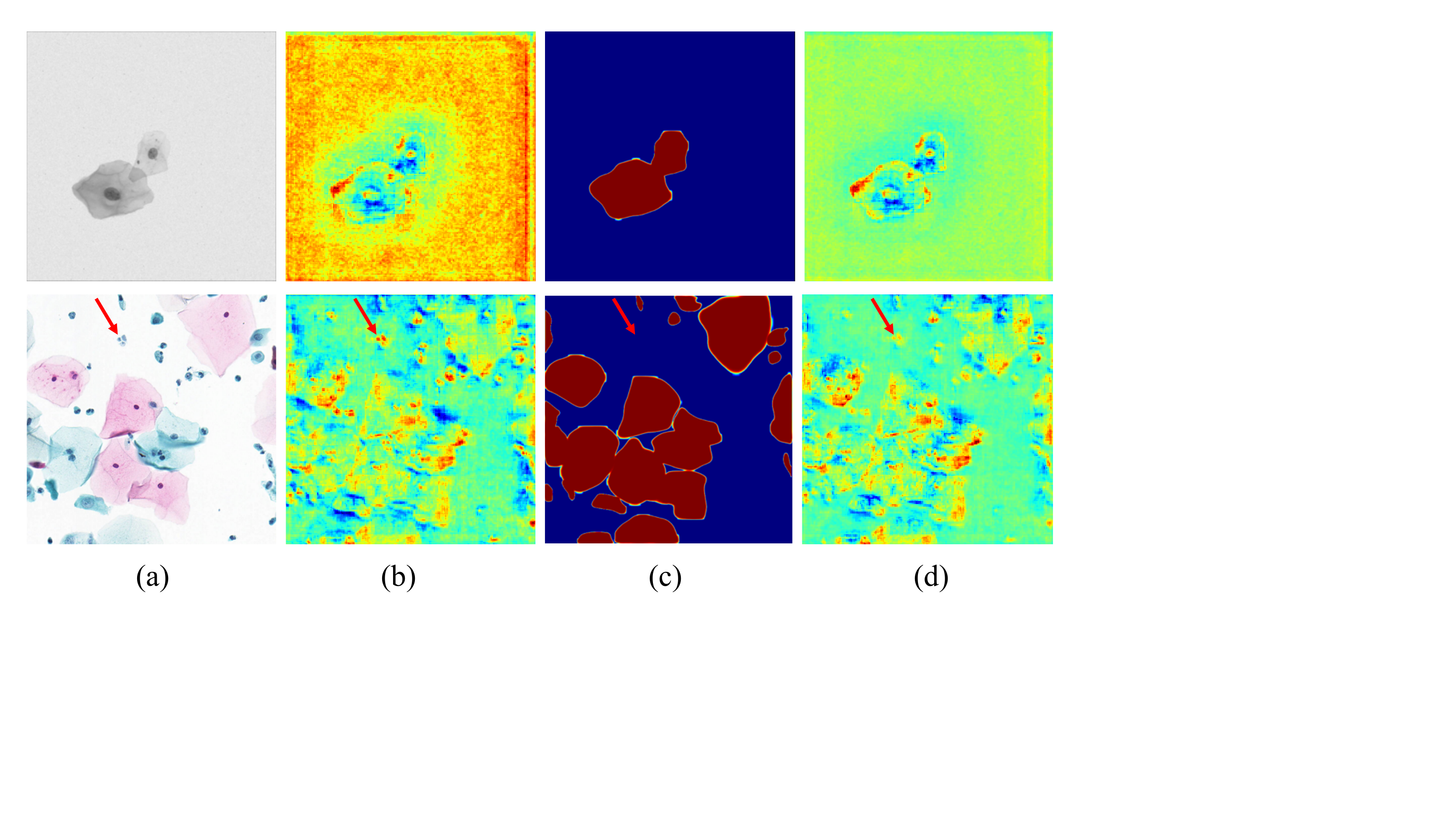}
         % \vspace{-0.1in}
	\caption{Illustration of background noise suppression in MRP: (a) the original image, (b) the origin feature map for region proposal, (c) the attention map from cytoplasm prediction, and (d) the re-weighted feature map for nuclei proposal generation.}
	%\label{fig:model}
	\label{fig:f-MRP}
	% \vspace{-0.2in}
\end{figure}

\input{Table/T3.tex}
\input{Table/T4.tex}

Applying MRP for mitigating the side effects from background mimics yields a further improvement of $0.59\%$ mAP and $0.31\%$ mAP on ISBI2014 and CPS datasets. Figure \ref{fig:f-MRP} provides the visualization of MRP operation, where background instances (e.g., mucus, karyoclasis, pointed by red arrow) are suppressed with strong responses in the feature map, encouraging the RPN to concentrate on cellular instance during nuclei region proposal.

\noindent \textbf{Design Choice for DRM:} 
We provide detailed comparisons on the ISBI2014 dataset to demonstrate the effectiveness of different components in DRM: 
1) $H_{i}$: instance mask head for coarse segmentation only; 2) $H_{o}$: intersection mask head for overlapping region segmentation; 
3) $H_{m}$: complement mask head for non-overlapping region segmentation; 

As seen in Table \ref{t4}, adding $H_{o}$ yields an improvement of $2.20\%$ in  average mAP, with a further $2.64\%$ gains from the additional $H_{m}$.
We notice that cytoplasm results have a more significant improvement of $13.2\%$ in mAP, which is indeed in line with our design intent.

\begin{figure}[!ht]
	\centering
	\includegraphics[width=0.8\linewidth]{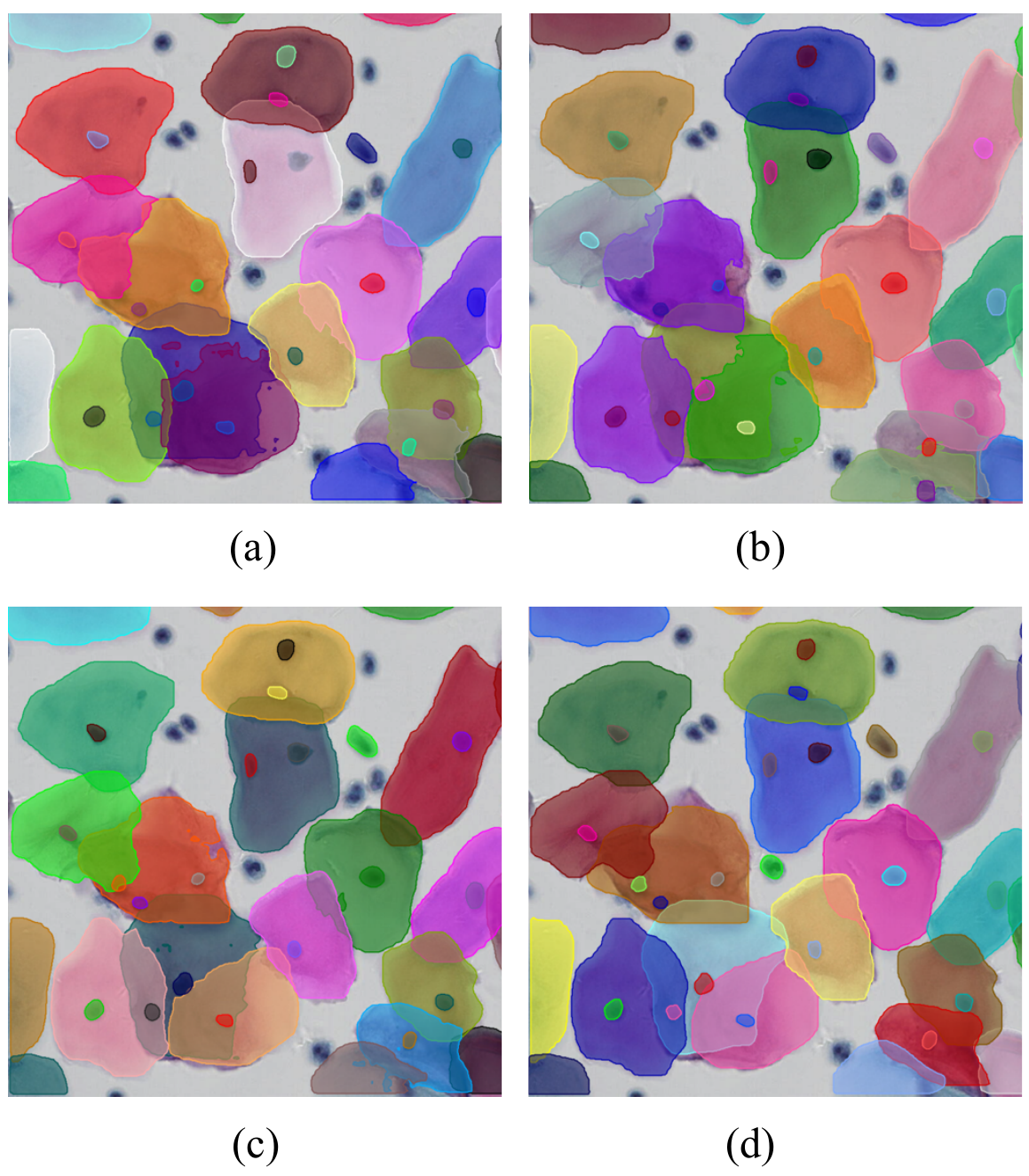}
	\caption{Qualitative results from (a) standard \cite{he2017mask}, (b) multi-task (DoNet w/o CRM), (c) amodal \cite{follmann2019learning}, and (d) proposed de-overlapping instance segmentation method. }
	\label{fig:f6}
\end{figure}

\noindent \textbf{Design Choice for CRM:} 
The goal of semantic consistency regularization is to 
enhance the model's overlapping reasoning capability by learning the concept of recombined instances from sub-regions.
We provide comparisons to demonstrate the design choice (Table \ref{t4}):
1) CU + FU: integration of rich semantic features as inputs via Concatenation Unit and Fusion Unit. 
2) $\mathcal{L}_{cons}$: consistency regularization between the recombined prediction $\hat{e}^{r}_{k,i}$ and the fusion of sub-regions $\hat{o}_{k,i}$, $\hat{m}_{k,i}$.
As seen in Table \ref{t4}, by aggregating the rich semantic feature for overlapping and non-overlapping region via Fusion Unit, we observe an increase of $1.52\%$ average mAP on the ISBI2014 dataset.
Adding consistency regularization further improves $0.91\%$ mAP for cytoplasm.

\begin{figure*}[!ht]
	\centering
	\includegraphics[width=\linewidth]{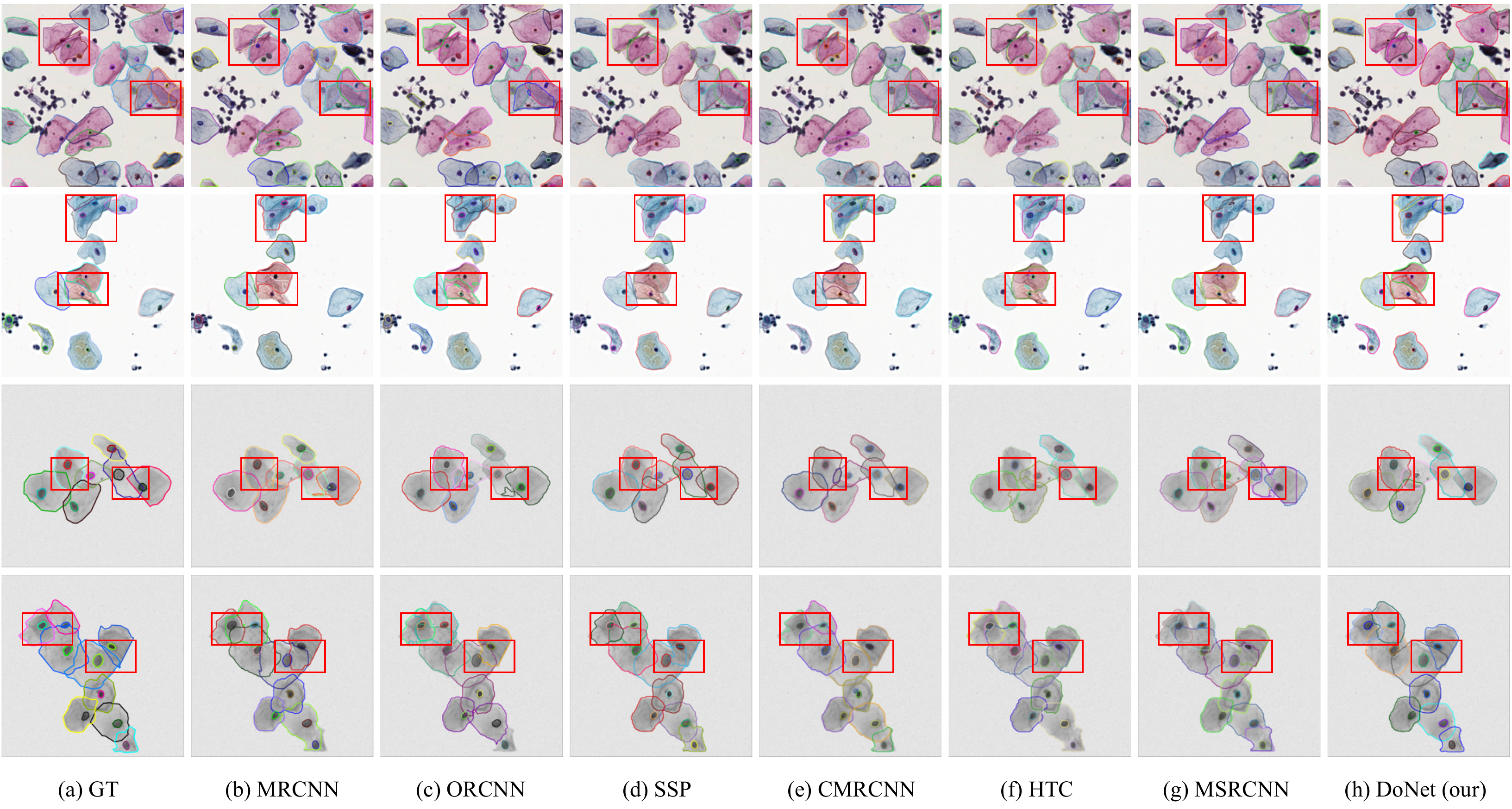}
	\vspace{-0.15in}
	\caption{Qualitative results of our DoNet and other SOTA methods on CPS (top) and ISBI2014 (bottom) datasets. (a) Ground Truth; (b) Mask R-CNN \cite{he2017mask}; (c) Occlusion R-CNN \cite{follmann2019learning}; (d) Xiao et al. \cite{xiao2021amodal}; (e) Cascade R-CNN \cite{cai2018cascade}; (f) Hybrid Task Cascade \cite{chen2019hybrid}; (g) Mask Scoring R-CNN \cite{huang2019mask}; (h) Our proposed DoNet.}
	\label{fig:f4}
  \vspace{-0.1in}
\end{figure*}
\begin{figure}[!ht]
	\centering
	\includegraphics[width=0.8\linewidth]{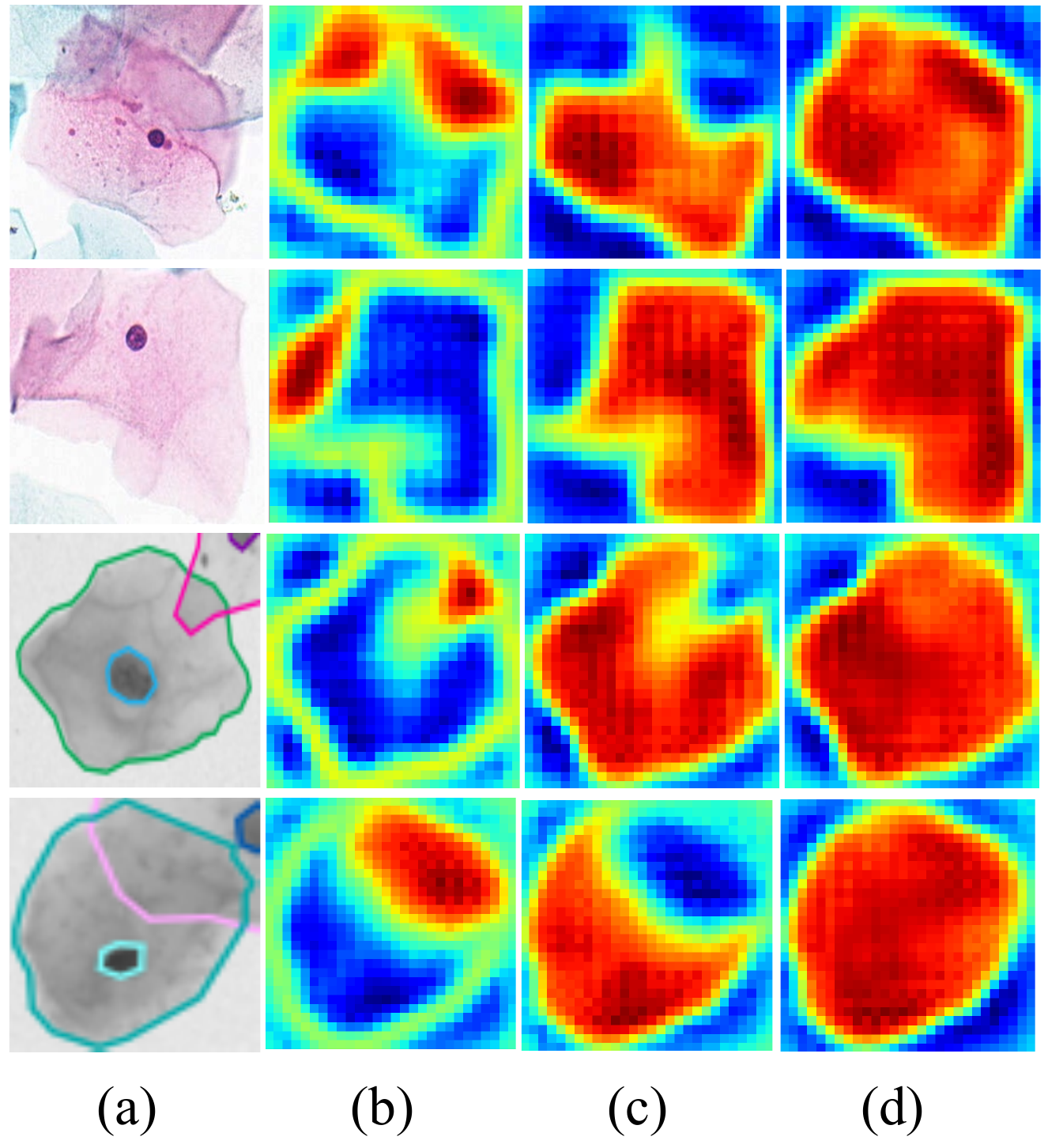}
	\vspace{-0.1in}
	\caption{Heatmaps of the intersection region, the
complement region, and the integral instance on CPS (top) and ISBI2014 (bottom) datasets, including (a) the original instance, the prediction heatmap of (b) intersection regions, (c) complement regions, and (d) the integral instance.}
	\label{fig:f3}
  \vspace{-0.2in}
\end{figure}

Figure \ref{fig:f6} visualizes the results of the DoNet and other typical instance segmentation methods, including standard (Mask R-CNN), multi-task (DoNet w/o CRM), and amodal (Occlusion R-CNN) instance segmentation model.
It demonstrates the importance of adding interaction among sub-regions via CRM and the strong perceptual capability of DoNet in overlapping regions. 

\subsection{Qualitative Analysis and Discussion}
We visualize the heatmap of the intersection region, the complement region, and the integral instance in Figure \ref{fig:f3}. The proposed method successfully identifies sub-regions based on the overlapping concept, even in low-resolution areas with high transparency.

Furthermore, we provide qualitative comparisons on CPS (top) and ISBI2014 (bottom) datasets in Figure \ref{fig:f4}. Our proposed DoNet outperforms other instance segmentation methods. Specifically, red rectangles provide details and highlight the main difference among these results. In the segmentation results of the CPS data, it can be seen that overlapped cells with different staining (e.g., dark red and blue) show significant appearance inconsistency between cellular sub-regions. Previous works (e.g., Mask R-CNN\cite{he2017mask} and Occlusion R-CNN\cite{follmann2019learning}) with limited perceptual capability have difficulty capturing the relationship between pixels in the intersection and complement sub-regions, leading to ambiguous segmentation contours. In contrast, our DoNet can effectively distinguish different instance boundaries and can better perceive the integrality of cells. This generalized superiority is also observed in the ISBI2014 dataset with low-contrast cellular instances.

%% file: Table/T1.tex
\begin{table*}[!ht]
	\caption{Quantitative segmentation results of DoNet and other state-of-the-art methods on CPS and ISBI2014.} 
	\vspace{-0.2in}
	\begin{center}{\small
			\resizebox{0.95\linewidth}{!}{
				\begin{tabular}{c|cccc|cccc}
					\hline
                     \multirow{2}{*}{Methods} & \multicolumn{4}{c|}{ISBI2014} & \multicolumn{4}{c}{CPS}\\
                    \cline{2-9} 
                    & mAP$\uparrow$&Dice$\uparrow$&F1$\uparrow$&AJI$\uparrow$&mAP$\uparrow$&Dice$\uparrow$&F1$\uparrow$&AJI$\uparrow$\\
                    \hline
	            Mask R-CNN \cite{he2017mask}&59.09 &91.15 &92.54 &77.07& 48.28 $\pm$ 3.10  & 89.32 $\pm$ 0.50 &85.07 $\pm$ 2.01 & 69.20 $\pm$ 2.27  \\
                    Cascade R-CNN\cite{cai2018cascade}&62.45 &91.29 &92.51 &77.91& 47.87 $\pm$ 3.27 & 89.24 $\pm$ 0.44 &83.33 $\pm$ 1.65 &68.86 $\pm$ 3.55  \\	
                    Mask Scoring R-CNN\cite{huang2019mask} &63.56 &91.28 &91.87 &75.14 & 48.38 $\pm$ 3.13 & 89.39 $\pm$ 0.24 &82.98 $\pm$ 1.86 &67.45 $\pm$ 2.45 \\	
                    HTC\cite{chen2019hybrid}&59.62 &91.39 &88.08 &75.00& 47.60 $\pm$ 3.56 &89.08  $\pm$ 0.51 &81.30 $\pm$ 2.56 &66.35 $\pm$ 2.84  \\	
                    Occlusion R-CNN\cite{follmann2019learning}&62.35 &91.75 &93.18 &78.64& 48.14 $\pm$ 2.84 & 89.08 $\pm$ 0.28 &85.69 $\pm$ 2.28 &69.51 $\pm$  2.45  \\	
                    Xiao et al.\cite{xiao2021amodal}&57.34 &91.70 &92.75 &78.29& 48.53 $\pm$ 2.85 & 89.29 $\pm$ 0.24 &85.46 $\pm$ 2.60 &69.37 $\pm$ 2.88  \\
                    DoNet &\textbf{64.02} &\textbf{92.13} &\textbf{93.23} &\textbf{79.05}& 49.43 $\pm$ 3.83 & \textbf{89.54 $\pm$ 0.25} &85.51 $\pm$ 2.33 &70.08 $\pm$ 2.84  \\	
                    DoNet w/ Aug.&- &- &-&-& \textbf{49.65  $\pm$ 3.52}  & 89.50  $\pm$ 0.38  &\textbf{86.30  $\pm$ 2.01}   &\textbf{70.56  $\pm$ 2.34}   \\	
					\hline
		\end{tabular}}}
	\end{center}
	\label{t1}
\end{table*}

%% file: Table/T2.tex
\begin{table}[!h]
\vspace{-0.1in}
	\caption{Comparison with other methods on ISBI2014.} 
 	\centering
			\resizebox{0.85\linewidth}{!}{
				\begin{tabular}{c|ccc}
					\hline
                     \multirow{2}{*}{Methods} & \multicolumn{3}{c}{ISBI2014} \\
                    \cline{2-4} 
                    & Dice$\uparrow$& $TP_{p}$$\uparrow$ & $FN_{o}$$\downarrow$ \\
                    \hline
                    Ushizima et al. \cite{ushizima2015segmentation}&0.872 &0.841&0.265 \\
                    Nosrati et al. \cite{nosrati2014variational}&0.871 &0.875&0.110 \\
                    Walter et al. \cite{walter2021multistar}&0.860 &0.830&0.310 \\
                    Lu et al. \cite{lu2015improved}&0.893 &0.905&0.315 \\
                    Lee et al. \cite{lee2016segmentation}& 0.897 & 0.882 &0.137   \\	
                    Tareef et al. \cite{tareef2018multi}& 0.898& 0.946 &0.161  \\	
                    Chen et al. \cite{chen2021segmentation}& 0.920& 0.900 &\textbf{0.020}  \\	
                    DoNet & \textbf{0.921}& \textbf{0.948} &0.162\\
					\hline
		\end{tabular}}
	\label{t2}
	\vspace{-0.1in}
\end{table}

%% file: Table/T3.tex
\begin{table*}[!ht]
\vspace{-0.1in}
	\caption{Effect of each proposed module on CPS and ISBI2014 datasets. \checkmark denotes adding the corresponding module.} 
 	\centering
			\resizebox{0.95\linewidth}{!}{
				\begin{tabular}{cccc|cccc|cccc}
					\hline
                        \multirow{2}{*}{Base} & \multirow{2}{*}{DRM} &\multirow{2}{*}{CRM} &\multirow{2}{*}{MRP}  & \multicolumn{4}{c|}{ISBI2014}& \multicolumn{4}{c}{CPS} \\
                    \cline{5-12} 
                    &&&& mAP$\uparrow$&Dice$\uparrow$&F1$\uparrow$&AJI$\uparrow$& mAP$\uparrow$&Dice$\uparrow$&F1$\uparrow$&AJI$\uparrow$\\
                    \hline
	              \checkmark&&&&59.09&91.15&92.54&77.07& 48.28 $\pm$ 3.10 & 89.32 $\pm$ 0.50 &85.07 $\pm$ 2.01 &69.20 $\pm$ 2.27   \\				
                    \checkmark&\checkmark&& &61.01&91.61&92.86&78.06& 48.03 $\pm$ 3.48   &89.13 $\pm$ 0.30 &84.63 $\pm$ 2.57 & 68.56 $\pm$ 2.57   \\
                     \checkmark&\checkmark&\checkmark&&63.43&91.87&\textbf{94.16}&\textbf{79.88}& 49.12  $\pm$ 3.26 &89.47  $\pm$ 0.31&84.82  $\pm$ 2.73 &69.26  $\pm$ 2.63\\
                    \checkmark&\checkmark&\checkmark&\checkmark&\textbf{64.02}&\textbf{92.13}&93.23&79.05&  \textbf{49.43 $\pm$ 3.83}  & \textbf{89.54 $\pm$ 0.25} &\textbf{85.51 $\pm$ 2.33 } &\textbf{70.08 $\pm$ 2.84 } \\	
					\hline
		\end{tabular}}
	\label{t3}
\end{table*}

%% file: Table/T4.tex
\begin{table*}[!ht]
	\caption{Ablation study of DRM and CRM on the ISBI2014 dataset.\checkmark denotes adding the corresponding component or strategy. } 
	\vspace{-0.15in}
	\begin{center}{\small
			\resizebox{\linewidth}{!}{
            \begin{tabular}{p{0.2cm}p{0.2cm}p{0.2cm}p{0.3cm}p{0.7cm}|ccc|ccc|ccc|ccc}
                \hline
                \multirow{2}{*}{$H_{i}$}&\multirow{2}{*}{$H_{o}$}&\multirow{2}{*}{$H_{m}$} &\multirow{2}{*}{FU}&\multirow{2}{*}{$\mathcal{L}_{cons}$}&\multicolumn{3}{|c|}{mAP$\uparrow$} & \multicolumn{3}{c|}{Dice$\uparrow$}& \multicolumn{3}{c|}{F1$\uparrow$}& \multicolumn{3}{c}{AJI$\uparrow$}\\
                \cline{6-17} 
                &&&&& Cyt.&Nuc.&Avg.& Cyt.&Nuc.&Avg.& Cyt.&Nuc.&Avg.& Cyt.&Nuc.&Avg.\\
                \hline
                \checkmark&&&&& 50.71 & 67.46 &59.09 & 90.72 & 91.59 &91.15& 86.96 & 98.13 &92.54&70.55 & 83.60 &77.07 \\
	        \checkmark&\checkmark&&&& 54.84 & 65.94 &60.39 & 91.56 & 91.54 &91.55& 87.00 & 97.68 &92.34& 72.07 & 82.67 &77.37 \\	
               \checkmark&\checkmark&\checkmark&&& 57.41 & 66.49 &61.95& 91.79 & 91.53 &91.66& 88.67 & 97.98 &93.32& 74.00 & 83.27 &78.63 \\
               \hline
               \checkmark&\checkmark&\checkmark&\checkmark&& 58.40 & \textbf{67.82} &63.11& \textbf{92.22} & \textbf{91.74} &\textbf{91.98}& 89.21 & 98.07 &93.64& 74.94 & 83.70 &79.32 \\
               \checkmark&\checkmark&\checkmark&\checkmark&\checkmark& \textbf{59.31} & 67.56 &\textbf{63.43}& 92.03 & 91.71 &91.87& \textbf{90.13} & \textbf{98.18 }&\textbf{94.16}& \textbf{75.86} & \textbf{83.91} &\textbf{79.88} \\
               \hline
		\end{tabular}}}
	\end{center}
	\label{t4}
\end{table*}

%% file: S5_Conclusion.tex
In this paper, we propose a de-overlapping network (DoNet) to address the overlapping challenges in cytology instance segmentation. The proposed DoNet enhances the model’s perception of overlapping regions by implicitly and explicitly modeling the interaction between cellular regions and the integral instance. Extensive experiments reveal the superiorities of the proposed DoNet, which not only provides immense potential for overlapping object perception in the medical domain but also occluded instance segmentation in general vision application scenarios.

%% file: S6_Acknowledgement.tex
This work was supported by National Natural Science Foundation of China (No. 62202403) and Beijing Institute of Collaborative Innovation Program (No. BICI22EG01).